\documentclass[runningheads]{llncs}

\usepackage{eccv}

\usepackage{eccvabbrv}

\usepackage{}
\usepackage{graphicx}
\usepackage{booktabs}
\usepackage{amsmath}

\DeclareMathOperator*{\argmin}{arg\,min}

\usepackage[accsupp]{axessibility}  

\usepackage{hyperref}

\usepackage{orcidlink}

\begin{document}

\title{RegionGrasp: A Novel Task for Contact Region Controllable Hand Grasp Generation} 


\titlerunning{RegionGrasp}

\author{Yilin Wang\inst{1} \and
Chuan Guo\inst{2}\orcidlink{0000-0002-4539-0634} \and
Li Cheng\inst{1}\orcidlink{0000-0003-3261-3533} \and Hai Jiang\inst{1}\orcidlink{0000-0003-1042-4897}}


\authorrunning{Wang et al.}

\institute{University of Alberta, Canada \and
Snap Research, USA}

\maketitle

\begin{abstract}
  Can machine automatically generate multiple distinct and natural hand grasps, given specific contact region of an object in 3D? This motivates us to consider a novel task of \textit{Region Controllable Hand Grasp Generation (RegionGrasp)},  as follows: given as input a 3D object, together with its specific surface area selected as the intended contact region, to generate a diverse set of plausible hand grasps of the object, where the thumb finger tip touches the object surface on the contact region. To address this task, RegionGrasp-CVAE is proposed, which consists of two main parts. First, to enable contact region-awareness, we propose ConditionNet as the condition encoder that includes in it a transformer-backboned object encoder, O-Enc; a pretraining strategy is adopted by O-Enc, where the point patches of object surface are randomly masked off and subsequently restored, to further capture surface geometric information of the object. Second, to realize interaction awareness, HOINet is introduced to encode hand-object interaction features by entangling high-level hand features with embedded object features through geometric-aware multi-head cross attention. Empirical evaluations demonstrate the effectiveness of our approach qualitatively and quantitatively where it is shown to compare favorably with respect to the state of the art methods.
\end{abstract}

\begin{figure*}
\begin{center}
    \includegraphics[width=0.85\linewidth]{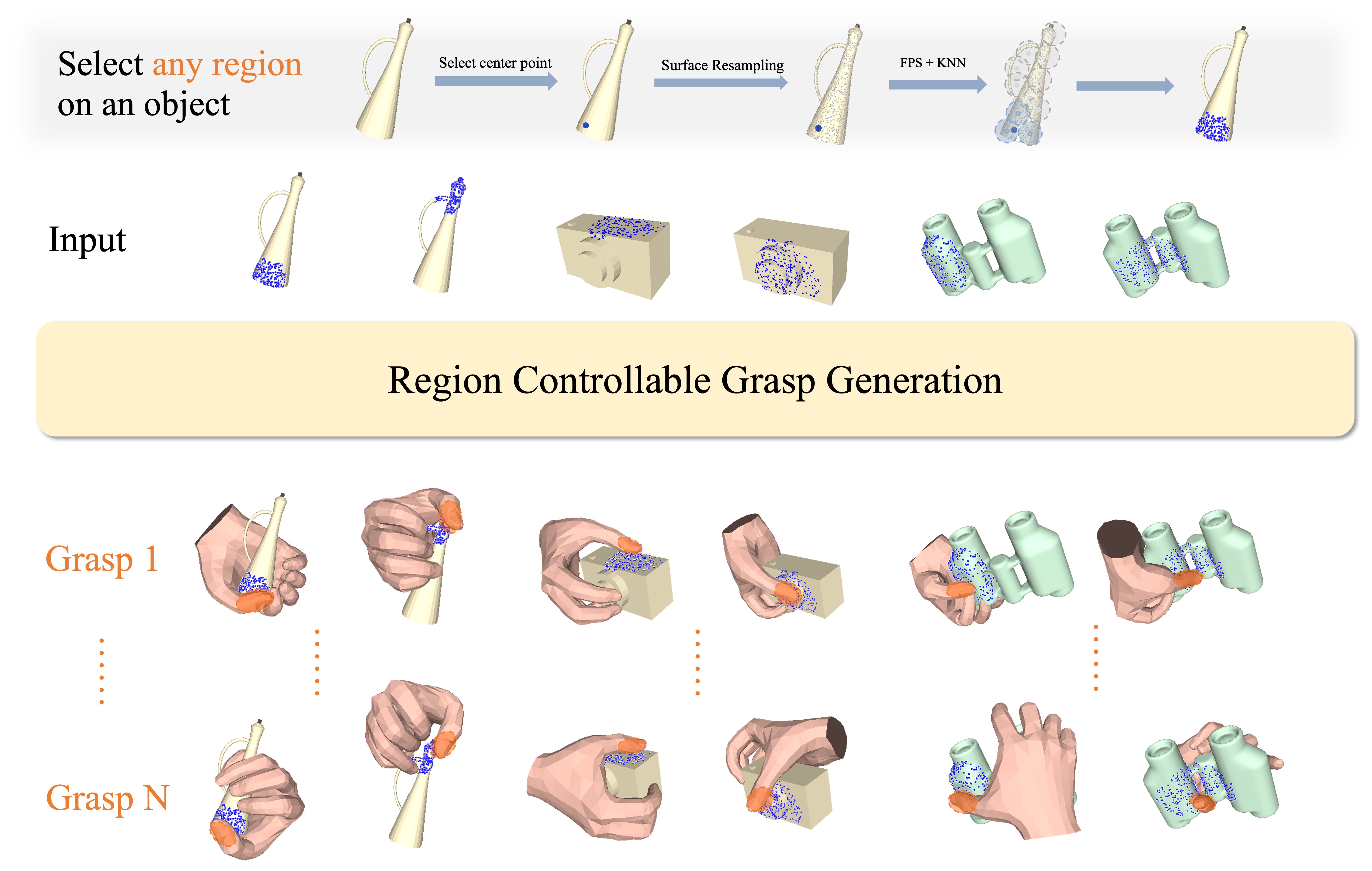}
\end{center}
   \caption{Illustration of the region controllable hand grasp generation task. Given an object with \textcolor{blue}{a specific condition region}, the task requires automatic generation of \textit{diverse} and \textit{natural} hand grasps with the  \textcolor{orange}{thumb finger tip} in contact with the condition region. Bottom 2 rows display the generated hand grasps from our proposed method RegionGrasp-CVAE,   presenting great diversity and controllability on both \textcolor{yellow}{in domain} and \textcolor{green}{out of domain} objects.}
\label{fig:overview}
\end{figure*}

\section{Introduction}
\label{sec:intro}

As a daily routine, we use our hands to grasp different kinds of objects for manipulating various tasks. Moreover, our dexterous hand finger configuration allows us to hold an object in different ways and perhaps for different purposes; even the same specific contact region of an object may often be grasped in multiple distinct manners. 
This raises the question: can a machine generate \textit{diverse} and \textit{natural} hand grasps that are in contact with prescribed surface regions of an object? 
There are existing works~\cite{GRAB, graspfield, graspTTA, eccv22graspD, contactopt, liu2023contactgen, ye2023ghop} considering the general hand grasp generation problem, yet they lack the level of controllability. Recent efforts\cite{cha2024text2hoi, christen2024diffh2o, chang2024text2grasp} on text-guided hand-object motion synthesis bring certain levels of temporal  and spatial controllability. Nevertheless, the intrinsic ambiguity and redundancy of high-level text prompts pose significant challenges for achieving precise spatial control in hand grasp synthesis, limiting their applicability in real-world scenarios like VR games, as in such applications, low-level control remains the preferred approach for accurately capturing diverse user intentions across different contexts.

In this work, we proposed a novel task named \textbf{Region Controllable Hand Grasp Generation(RegionGrasp)} to bridge this gap, as shown in Figure \ref{fig:overview}. Provided as input a 3D object with its specific surface area selected as the intended contact region, our goal is to generate a diverse set of plausible hand grasps of the object,
where the thumb finger tip touches the object surface on
the intended contact region; the thumb finger tip is specially designated as it is the most frequently involved region in the interaction with objects. In particular, two issues remain to be addressed in this task. The first is \textbf{region-awareness}: \textit{where} (geometrically) and \textit{what} (semantically) the contact region locates on the object surface. The second is \textbf{interaction-awareness}: how to represent the interactions between hand and 3D object along the contact surface region.

The aforementioned challenges lead us to propose \textbf{RegionGrasp-CVAE}, a conditional variational autoencoder (CVAE~\cite{CVAE}) based generative model. The input of our model is the resampled point cloud of the given 3D object mesh, with a point representing the center of the intended condition region, as shown in the first row of Figure \ref{fig:overview}. 
The output is a possible hand grasp mesh parameterized by MANO model~\cite{MANO:SIGGRAPHASIA:2017}. 

To enable region-awareness, we use ConditionNet as the condition encoder, consisting of a transformer-backboned O-Enc, and a MLP-based condition region encoder; the O-Enc serves to embed spatial and geometric features of the object point cloud and the selected region. To strengthen the capability of capturing geometric \& semantic information in O-Enc, a pre-training strategy is further adopted, where point patches of objects are randomly masked off, then restored by the O-Enc. In the meantime, the HOINet is introduced to realize interaction-awareness: it contains a transformer-backboned H-Enc to embed hand features; this is followed by a HOI encoder, designed to embed hand-object interaction (HOI) features by entangling high-level hand and object features through geometric-aware multi-head cross attention blocks. Overall, our framework shows the benefits of stable and effective learning with simple training objectives by engaging merely standard VAE reconstruction loss and KL-Divergence constraint, involving little human intervene. In other words, it is not necessary in our approach to rely on those commonly used interaction annotations and post refinement as in~\cite{GRAB, graspfield, graspTTA, contact2grasp, liu2023contactgen}.

Our main contributions can be summarized as follows:
1) A new task of region-based controllable hand grasp generation was conceived in our work, in which generated hand grasps are expected to contact with a selected surface region of an object by the thumb finger tip. 
2) In addressing the new task, we proposed RegionGrasp-CVAE, a simple yet effective baseline model based on CVAE framework, where a ConditionNet with pre-trained strategy is utilized to realize region-awareness, and an HOINet component to account for interaction-awareness. Empirical evaluations demonstrate qualitatively and quantitatively the effectiveness of our approach. Codes will be released on \url{https://github.com/10cat/RegionGrasp}.

\begin{figure*}
\begin{center}
    \includegraphics[width=1.0\linewidth]{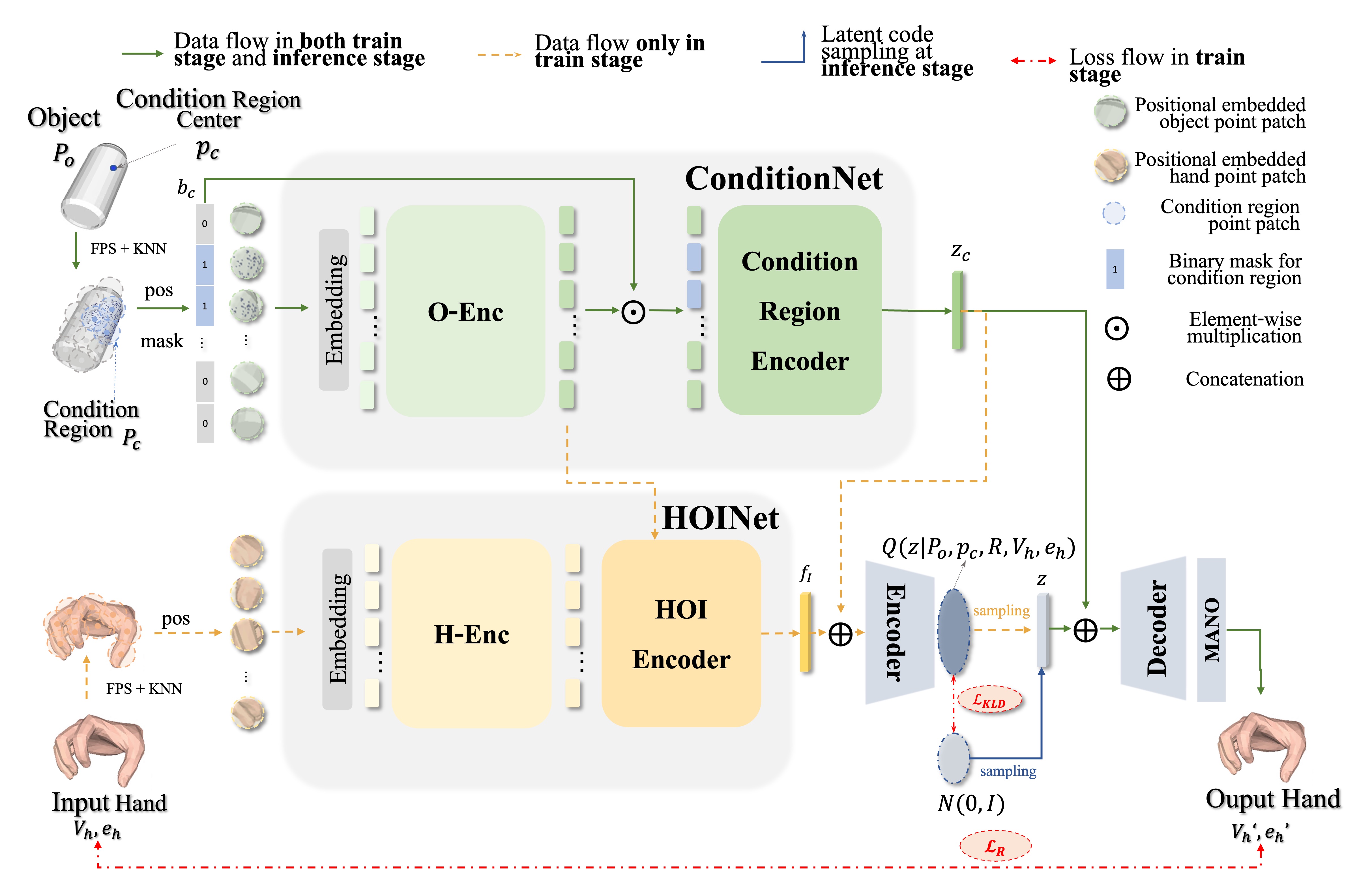}
\end{center}
   \caption{\textbf{An overview of our RegionGrasp-CVAE framework with training and inference pipeline}. The input object are resampled to point cloud and grouped into point patches. The binary condition region mask is generated based on the point patches to distinguish the condition region from other object regions. \textcolor{green}{The ConditionNet} embeds geometric-aware object tokens through O-Enc, which are then partially masked by condition region mask and finally encoded as a global region-aware feature vector $z_c$ by Condition Region Encoder. \textbf{During training}, hand tokens embedded by H-Enc interact with object tokens from O-Enc through \textcolor{yellow}{HOI encoder} to encode hand-object interaction(HOI) features $f_I$ that yields the posterior distribution $Q(z|P_o, p_c, R, V_h, e_h)$. A latent vector is sampled from the posterior distribution, concatenated with $z_c$, and then mapped back to the hand mesh space through VAE decoder and MANO model. \textbf{During inference}, the latent code $z$ is randomly sampled from standard Gaussian distribution. The VAE-Decoder takes the concatenation of $z_c$ with a sampled latent code vector $z$ to generate MANO hand parameters which are then regressed to the output hand shape.}
\label{fig:model}
\end{figure*}

\section{Related Work}
\label{sec:realted}
\subsection{Hand Grasp Generation}
Grasp synthesis has been a longstanding problem with a series of in-depth studies leading to significant advancements in both robotic grasp synthesis\cite{lundell2021multi, cvpr22dgrasp, xu2023unidexgrasp, turpin2023fastgraspd, xu2024dgtr, li2024graspmulti} and human hand grasp synthesis\cite{GRAB, graspfield, graspTTA, eccv22graspD,contact2grasp, contactopt, zhou2022toch, liu2023contactgen, ye2023ghop, cha2024text2hoi, christen2024diffh2o, chang2024text2grasp}; in this work, we primarily focus on the latter. A majority of previous efforts have been devoted to learning a more expressive and plausible hand-object contact prior, grounded on either CVAE-based\cite{GRAB,graspfield,graspTTA,contact2grasp,liu2023contactgen}, denoising-based\cite{zhou2022toch, contactopt, ye2023ghop} or physics-based\cite{cvpr22dgrasp, eccv22graspD} framework.  \cite{GRAB} proposed to utilize signed distance map for pose refinement. \cite{graspTTA} learns a contact map simultaneously during training, which is applied as the pseudo label for test-time optimization. In \cite{liu2023contactgen}, contact map, part map and direction map are learned hierarchically through CVAE to embed hand-object contact semantics. \cite{ye2023ghop} model a diffusion-based generative prior utilizing text annotations. While achieving realistic and plausible synthesis of hand grasps given different kinds of object, these methods lack controllability over generated hand grasps, limiting their practical application. 

Recent works\cite{cha2024text2hoi, christen2024diffh2o} have aimed to address this by generating temporally controllable hand-object motions guided by text, yet they fail to provide any spatial control over hand-object contact. The only effort so far on spatial controllability of hand grasp synthesis model\cite{chang2024text2grasp} introduced LLM-generated text prompts to describe object contact parts, which is employed as the control signal over hand-object contact region. However,the inherent vagueness and redundancy of high-level text prompts still leaves a huge gap to executing precise spatial controllability of hand grasp synthesis, which  makes it hard to get generalized in real-world applications such as VR games, where low-level control is still often the choice for accurate description of various user intentions in all kinds of scenarios. To address this, our method contribute to modelling contact condition region with an explicit geometric representation based on point patches, enabling low-level spatial controllability through learning a geometric-aware conditional generative prior.


\subsection{Deep Learning of 3D Point Cloud}
3D point cloud, as one of the most accessible 3D representation, has motivated a large number of recent research  activities \cite{pointnet, pointnet++, pointtrans, pointbert, pointmae, yu2021pointr, yuan2018pcn} to seek effective deep learning solutions to various downstream tasks. The PointNet and related ideas \cite{pointnet, pointnet++} have enabled the foundation of CNN-based architecture and grouping strategies. \cite{pointtrans} incorporated the transformer architecture to process point cloud data via a dedicated transformer block. In \cite{pointmae}, a self-supervised pretraining strategy is used based on mask-autoencoding, without utilizing any other pseudo labels. A geometric-aware transformer block is conceived by the work of \cite{yu2021pointr}. Our approach has adopted the pretraining strategy from \cite{pointmae} to enhance the object condition embedding. Also the architecture of our HOI Encoder is inspired by \cite{yu2021pointr}.

\section{Methodology}
\label{sec:method}
\begin{figure}[ht]
\centering
\setlength{\tabcolsep}{5pt} 
\begin{tabular}{cc}
    \begin{minipage}{0.4\textwidth}
        \centering
        \includegraphics[width=\linewidth]{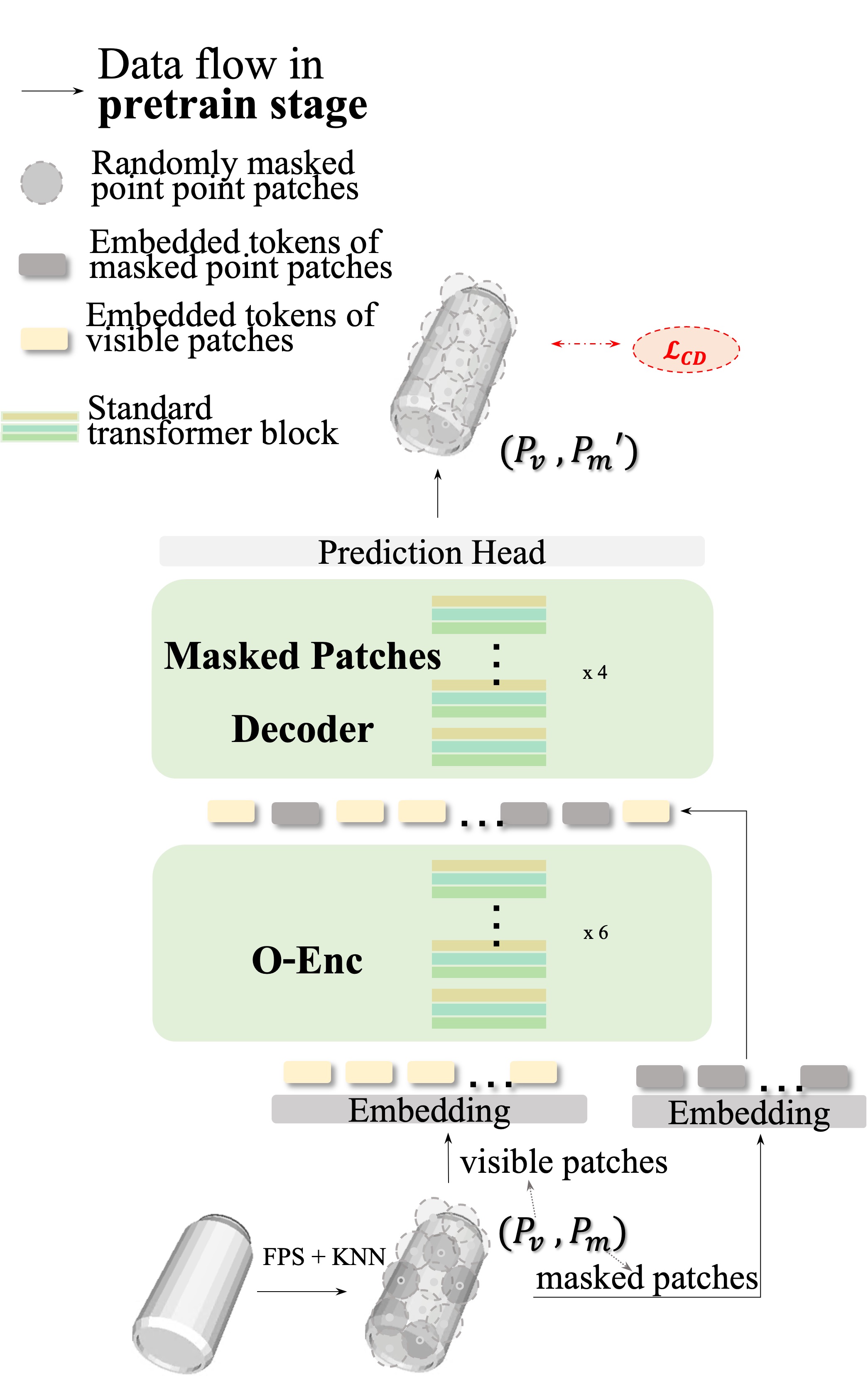}
        (a)
    \end{minipage} &
    \begin{minipage}{0.6\textwidth}
        \centering
        \includegraphics[width=\linewidth]{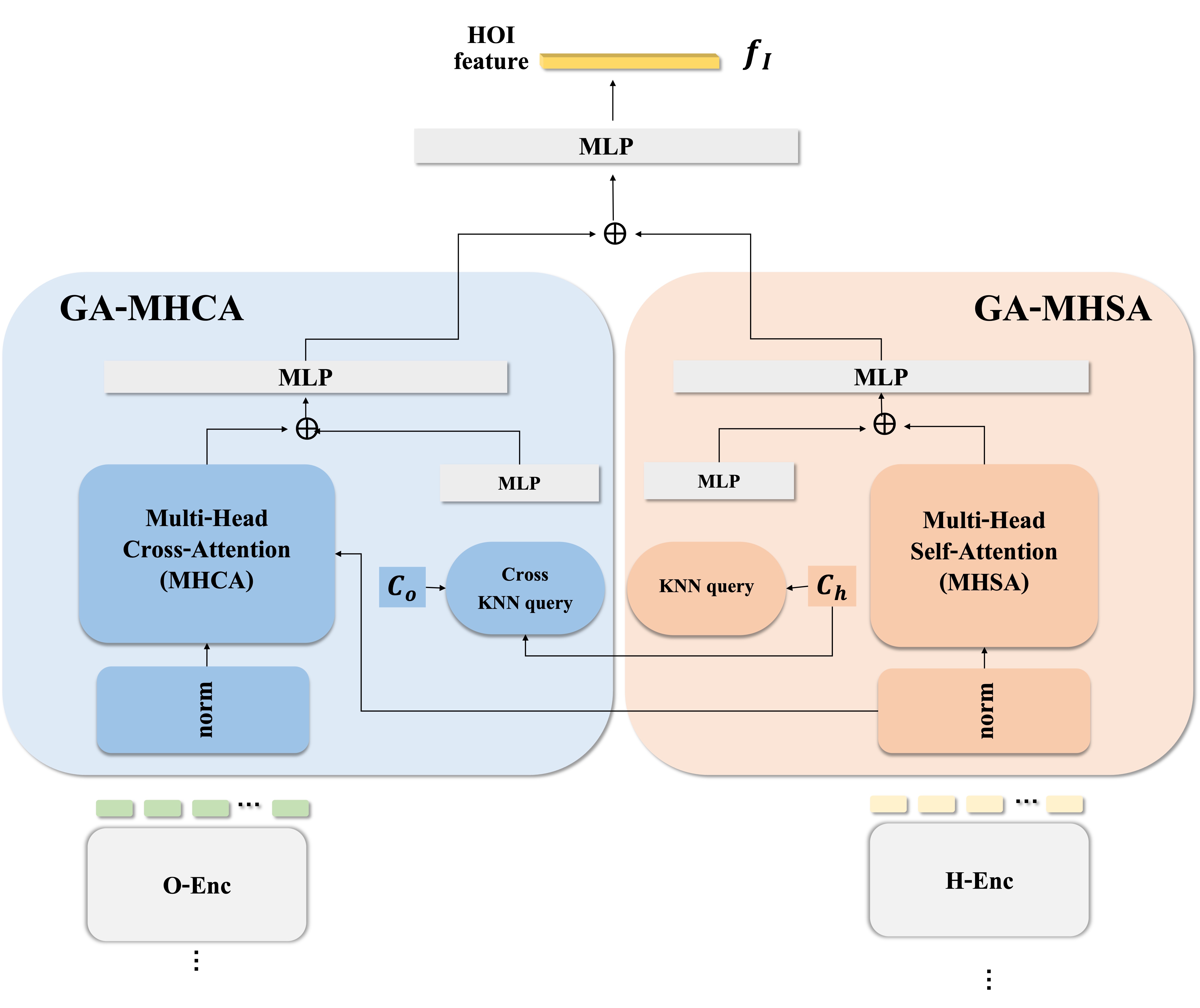}
        (b)
    \end{minipage}
\end{tabular}
\caption{\textbf{An overview of our RegionGrasp-CVAE framework}. (a) Pretrain pipeline for O-Enc based on mask auto-encoding. (b) Elaborated architecture of GA-MHSA and GA-MHCA blocks in the HOI Encoder, the core component of our HOINet.}
\label{fig:details}
\end{figure}

Our method aims to approach the region controllable hand grasp generation task by training a CVAE-based generative model, namely RegionGrasp-CVAE. The proposed architecture exploits the strength of the CVAE framework and transformer backbone, as shown in Figure \ref{fig:model}. It was generally established upon the observation that region-awareness and interaction-awareness are the two principle challenges in our proposed task. To enable the region-awareness, specifically \textit{where} and \textit{what} the condition region is, it is crucial to capture low-level geometric features (\textit{where}) of the given object, as well as the high-level semantic features (\textit{what}) of the condition region. For this reason, we established ConditionNet which consists of a transformer-based object encoder O-Enc, and a Condition Region Encoder; the former extracts low-level geometric features, while the latter models high-level semantic information. To further enhance the low-level geometric modeling, a pretrain strategy was introduced based on mask auto-encoding, elaborated in Figure \ref{fig:details}(a) and Section \ref{sec:pretrain}.

Interaction-awareness particularly emphasizes on how the hand is deformed to grasp the target region. This leads to the proposed HOINet, which incorporates the feeds from both the hand encoder H-Enc and object encoder O-Enc to yield high-level interaction features, through dedicated Geometric-Aware Multi-head Self-Attention (GA-MHSA) and Geometric-Aware Multi-head Cross-Attention (GA-MHCA), elaborated in Figure \ref{fig:details}(b) and Section \ref{sec:HOI}.

\textbf{At the training stage}, the input of the RegionGrasp-CVAE includes the resampled object point cloud $P_o \in \mathbb{R}^{N \times 3}$, the condition region center $p_c \in \mathbb{R}^{3}$ with region size $R \in \mathbb{R}$, and hand vertices point cloud $V_h \in \mathbb{R}^{778 \times 3}$ of the input hand mesh $(V_h, e_h)$. Point clouds of hand and object, as well as the condition region are reformed into point patches representation. 
The transformer-based ConditionNet embeds the region-aware object condition feature as the condition vector $z_c \in \mathbb{R}^{d_c}$. HOINet embeds and then entangles the hand features with the object features, resulting in $f_I \in \mathbb{R}^{d_h}$. The concatenated vector of $f_I$ and $z_c$ yields the posterior distribution $Q(z|P_o, p_c, R, V_h, e_h)$ through VAE-encoder. A latent vector $z$ is sampled from the posterior distribution, concatenated with $z_c$, and then mapped back to the MANO parameter space through VAE-decoder. The following MANO layer is learned to fit MANO parameter to the reconstructed hand mesh $(V_h', e_h')$. We optimized the model by simultaneously minimizing the reconstruction loss between $(V_h', e_h')$ and $(V_h, e_h)$, as well as the KL-divergence of constraining the posterior distribution to a standard Gaussian distribution.

\textbf{At the inference stage}, a condition vector is extracted from our ConditionNet given an input object point cloud with a condition region. Given an object point cloud with a selected contact region, the condition vector is extracted from ConditionNet and concatenated with a latent code randomly sampled from the standard Gaussian distribution. By taking the resulted conditional latent vector, the optimized VAE-decoder and MANO layer decodes the generated hand mesh as output. See supp. for implementation details.

\subsection{Representation based on Point Patches \label{sec:patch}}
To process the hand/object point cloud more effectively and efficiently, we grouped the point cloud into point patches following the procedure in \cite{pointbert}. Given an input object point cloud $ P_o \in \mathbb{R}^{N \times 3}$, center-points $C_o \in \mathbb{R}^{G_o\times 3}$ are selected using farthest point sampling (FPS)\cite{pointnet++}. The point cloud is then grouped into point patches $\{p_o^{(i)}\}^{G_o}_{i=1}$, where $p_o^{(i)} \in \mathbb{R}^{S\times 3}$ is a subset of $P_o$ consisting of the nearest neighbors of center-point $C_o^{(i)}$ obtained through computing K-Nearest Neighbors (KNN) with $k=S$. Each point patch is normalized to the center-point coordinate. Similarly, center points and point patches for the hand can be obtained as $C_h \in \mathbb{R}^{G_h\times 3}$ and $\{p_h^{(i)}\}^{G_h}_{i=1}$.

To form a reasonable and configurable condition region representation, we characterized the condition region by a selected center point $p_c \in \mathbb{R}^3$ of the region. Top-$R$ nearest object point patches are selected as the representation of the specific condition region on the given object $P_{c} \in \mathbb{R}^{RS \times 3}$, with $R$ indicating the size of the region. The condition region binary mask is defined as $b_c = [c_0, c_1, ..., c_{G_o-1}]$, where $c_i = 1$ if $ p^{(i)}_o \in P_c$ otherwise $c_i = 0$.



\subsection{ConditionNet}
Our proposed ConditionNet consists of two major modules. The O-Enc is backboned on standard transformer blocks, which takes the embedded tokens of all the object point patches and output the tokens enhanced with geometric and semantic features of the object. The enhanced tokens are then element-wisely multiplied with the condition region binary mask, and processed by the Condition Region Encoder to obtain the region-aware object global feature as the condition vector $z_c$. 


\textbf{Pretrainig O-Enc.\label{sec:pretrain}} To generate plausible hand grasps given the certain object with the specific target region to grasp, we argue that it is important to optimize the condition embedding module for extracting effective features that can well characterize the geometric information of the given object. We introduced a pretrain strategy based on mask autoencoder framework \cite{pointmae} to enhance the geometric-awareness of our O-Enc, as shown in Figure \ref{fig:details}(a).  During pretraining, certain proportion of the object point patches is randomly masked off, denoted as $P_m$. The O-Enc takes only the embedded tokens of unmasked point patches $P_v$. A similar but shallower transformer decoder processes the combined tokens of both $P_m$ and $P_v$ for predicting the coordinates of the masked point patches $P'_m$.



\subsection{HOINet \label{sec:HOI}}
Instead of relying on extra interaction-related annotations for imposing the model to embed hand-object interaction features \cite{graspfield, graspTTA, GRAB}, we realized interaction-awareness by proposing HOINet to 1) embed the hand features with H-Enc and 2) entangle the semantic and spatial features of hand and object to embed hand-object interaction features to learn better latent code mapping, using HOI Encoder backboned on geometric-aware transformer blocks.

Our proposed H-Enc and the embedding module ahead share similar structures respectively with our O-Enc and its embedding layers, except that less transformer blocks are used and the point patch configuration differs. 

\textbf{Interaction-aware HOI Encoder.} We proposed an interaction-aware HOI Encoder to embed hand-object interaction features into embedded hand tokens, shown in Figure \ref{fig:details}(b). As hand-object interaction features are strongly related with the spatial relationship between hand and object, we established Geometric-Aware Multi-head Self-Attention (GA-MHSA) and Geometric-Aware Multi-head Cross-Attention (GA-MHCA) based on geometric-aware transformer blocks \cite{yu2021pointr}. KNN query embedding in GA-MHSA directly encodes the spatial features of the hand point cloud by embedding top-k nearest neighbors for every $c\in C_h$, while Cross KNN query in GA-MHCA retrieves top-k candidates for every $c\in C_o$ from $C_h$. Both GA-MHSA and GA-MHCA concatenate the semantic features embedded from the MHCA/MHSA module with the spatial features. The MHCA module takes the tokens from both O-Enc and H-Enc, while the MHSA module takes only the hand tokens from H-Enc.






\subsection{Loss Function}
In the pretrain stage, we applied Chamfer Distance loss to optimize O-Enc and Masked Patches Decoder. With the $N_m$ and $N'_m$ denoting the number of points in $P_m$ and $P'_m$ respectively:
\begin{equation}
    \mathcal{L}_{pretrain} = \frac{1}{N_m}\sum\limits_{x\in P_m} \min\limits_{y\in P'_m}\parallel x- y \parallel^2_2 + \frac{1}{N'_m}\sum\limits_{y\in P'_m} \min\limits_{x\in P_M}\parallel x- y \parallel^2_2
\end{equation}

In the training stage, as mentioned above our method only use the hand mesh reconstruction loss and standard KL-Divergence loss for training. The hand mesh reconstruction loss is a combination of hand vertices loss and hand mesh edges loss, where we applied L1 norm for both, where $\mathcal{L}_{verts} = \parallel V_h' - V_h \parallel_1,\quad\mathcal{L}_{edges} = \parallel e_h' - e_h \parallel_1$.Following standard VAE \cite{CVAE}, we use KL-Divergence loss to enforce the latent code distribution to be close with the standard Gaussian distribution of the same dimensionality, formulated as $\mathcal{L}_{KLD} = \mathrm{KL}(Q(z | P_o, p_c, R, V_h, o_h)\parallel \mathcal{N}(0, I))$.

The total loss we applied for training can hence be formulated as the combination of all the loss above as
\begin{equation}
    \mathcal{L}_{train} = \lambda_v\mathcal{L}_{verts} + \lambda_e\mathcal{L}_{edges} + \lambda_{KLD}\mathcal{L}_{KLD}
\end{equation}

\section{Experiment}

\subsection{Implementation Details}
We configure the point patch number $G=128$ and size $S=32$ for both hand and object. The object vertices number used for training and testing is $N =2048$. The size of the condition region $R = 16$ is equal to 12.5\% of all the point patches. For pretrain strategy based on random masking, we choose masking ratio equal to 60\%. We use AdamW optimizer for the transformer-based modules with initial learning rate $lr=$5e-4, and CosLR scheduler applied. More implementation details to be found in the supplementary materials.

\subsection{Datasets}
We explored two hand-object datasets for training and evaluation.

\textbf{ObMan dataset\cite{hasson19_obman}} is a popular synthetic dataset of hand-object interaction with 2772 house-hold objects of 8 different categories from ShapeNet \cite{shapenet} dataset. The ground truth hand grasps in the dataset are generated by GraspIt software. ObMan is used to train and evaluate our model. The trainset objects are also used as part of our pretrain dataset.

\textbf{GRAB dataset\cite{GRAB}} is a real human-object interaction dataset consisting of full 3D body shape and pose sequences of 10 subjects interacting with 51 everyday objects. Our evaluation on GRAB\cite{GRAB} only involves the subset with only the hand part interacting with objects. Note that although the amount of objects is far less than that of ObMan, we observe that the objects in GRAB\cite{GRAB} cover more diverse categories and sophisticated structures. We use the objects in trainset to expand our pretraining dataset, and the testset for evaluation of generalization ability.

\subsection{Evaluation Metrics}
We implemented several quantitative measures to assess the quality and diversity of our region-based conditional hand grasp generation task.

\textbf{Condition Hit Rate (CR, $\%$)}: We proposed the Condition Hit in Region (CR) Rate to assess the controllability with respect to the condition region. Denoting the thumb pulp vertices of the predicted grasp as $P_{thb}$, the CR rate is defined as $CR = \frac{\text{card}\{P_{CR}\}}{\text{card}\{P_{o2t}\}}$, where $P_{o2t} = \bigcup\limits_{q\in P_{thb}} \argmin\limits_{p \in P_o} \lVert q - p \rVert^2$, and $P_o$ represents the object point cloud. The subset $P_{CR}$ is selected such that $P_{CR}= P_c \cap P_{o2t}$, where $P_c$ denotes the specified condition region.

\textbf{Condition Contact - Interpenetration Ratio (CCA / IV, $cm^{-1}$)}: Previous works often measure the contact area (CA) to evaluate the efficacy of generated hand grasps in maintaining contact with objects, and interpenetration volume (IV) to gauge the degree of interpenetration of generated hand meshes with object meshes. However, reporting CA or IV independently is not intuitive for assessing the quality of generated grasps due to the inherent strong trade-off between these metrics. Inspired by \cite{eccv22graspD}, we report the ratio of conditional contact area (CCA) to IV, defined as $\frac{CCA}{IV} = \frac{CR \times CA}{IV}$, to better reflect the overall quality of generated hand grasps in our region-specific scenario. Here CR, CA and IV all refer to the mean statistics across all generations in the testset.


\textbf{Grasp Displacement Ratio (GDR)}: Grasp displacement (GD) is a commonly used physic-based simulation metric for evaluating the stability of a grasp \cite{hasson19_obman, graspfield, graspTTA, eccv22graspD}. By placing generated grasps and corresponding objects into a physics simulator with a fixed hand position, we measure the displacement distance of the object under gravity for the generated grasp. Here we use the ratio of GD (GDR) of generated hand grasps to that of the ground truth as a normalized scale for this metric.

\textbf{Diversity (DivDist, $mm$)}: To assess the generative diversity of our model, we introduce diversity metrics following \cite{Guo_2022_CVPR}. For each object and condition region, we generate 20 distinct samples and randomly divide them into two groups of 10. The mean square error (MSE) between the samples of the two groups is reported as the diversity metric.


\begin{table}[tb]
  \caption{Quantitative results on ObMan\cite{hasson19_obman} and GRAB\cite{GRAB}.
  }
  \label{tab:SOTA}
  \centering
  \begin{tabular}{@{}cccccc@{}}
    \toprule
    Method & Test dataset & CR Rate (\%)$\uparrow$ & CCA/IV$\uparrow$ & GDR$\downarrow$ & DivDist (mm)$\uparrow$ \\
    \midrule
    GrabNet\cite{GRAB} & ObMan & 10.24 & 0.80 & \textbf{0.89} & 0.64 \\
    GraspTTA\cite{graspTTA} & ObMan & {15.38} & 1.31 & 0.91 & 0.69 \\
    Contact2Grasp\cite{contact2grasp} & ObMan & 10.87 & 1.28 & 1.01 & 0.57 \\
    ContactGen\cite{liu2023contactgen} & ObMan & 12.29 & 1.12 & 1.05 & 1.29 \\
    Ours & ObMan & \textbf{87.88} & \textbf{1.63}& 1.10 & \textbf{2.48} \\
    \hline
    GraspTTA\cite{graspTTA} & GRAB & 4.34 & 0.01 & 10.68 & 0.67 \\
    Ours & GRAB & \textbf{85.76} & \textbf{1.42} & \textbf{4.28} & \textbf{2.32} \\
  \bottomrule
  \end{tabular}
\end{table}

\begin{figure*}
\begin{center}
    \includegraphics[width=1\linewidth]{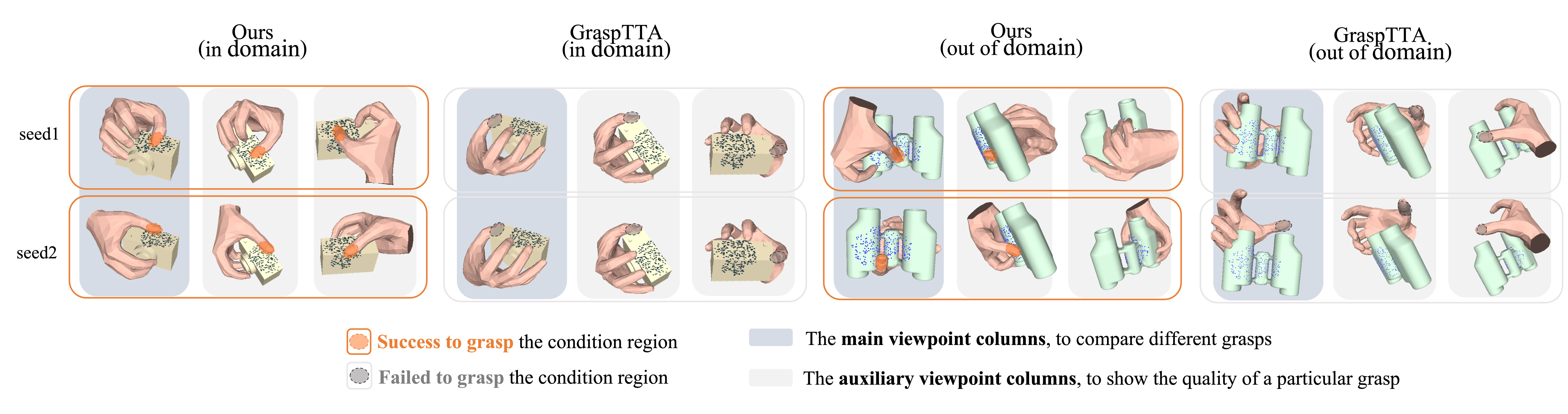}
\end{center}
   \caption{Qualitative comparison with GraspTTA\cite{graspTTA}.The best 2 grasps are selected for each method tested on in-domain / out-of-domain objects.}
\label{fig:compare}
\end{figure*}

\subsection{Results}
Here we present our quantitative and qualitative results. Note that CR Rate is reported by taking the average of 5 groups of different set of condition regions.



\subsubsection{Quantitative Results}  Table \ref{tab:SOTA} shows the comparison results on ObMan testset, where we mark \textbf{the best} for each metric. To introduce the same contact region condition for fair comparison, we employ our Condition Region Encoder in revising the object encoder of \cite{GRAB, graspTTA, contact2grasp}, and utilize the condition region mask for obtaining an condition map additional to the contact map, part map and direction map in \cite{liu2023contactgen}. As presented, our method shows great competitive on the contact quality and grasp stability of generated grasps compared to the state of the art approaches that are facilitated with test time optimization \cite{GRAB, graspTTA, contact2grasp, liu2023contactgen}. On generation diversity and CR Rate measures, our method largely outperforms the state of the art methods. The results have well demonstrated the validity of our proposed task, as the low CR Rate and low diversity measures of previous methods show the poor ability of adapting generated grasps to different regions of the objects. In contrast, our methods show promising results on performing diverse and plausible hand grasp generation targeted on specific region of the given object.

We also tested the generalization ability of our methods on the testset of GRAB\cite{GRAB} dataset and compared with the results of \cite{graspTTA} which is obtained under the same experiment setting. As shown in last two rows in Table \ref{tab:SOTA}, our methods can be well generalized to the out of domain objects with reasonable grasp quality and conditioned generation performance preserved, while \cite{graspTTA} fails to generate reasonable grasps without high-quality labels for interaction.


\subsubsection{Qualitative Results} 
\begin{figure*}
\begin{center}
    \includegraphics[width=1\linewidth]{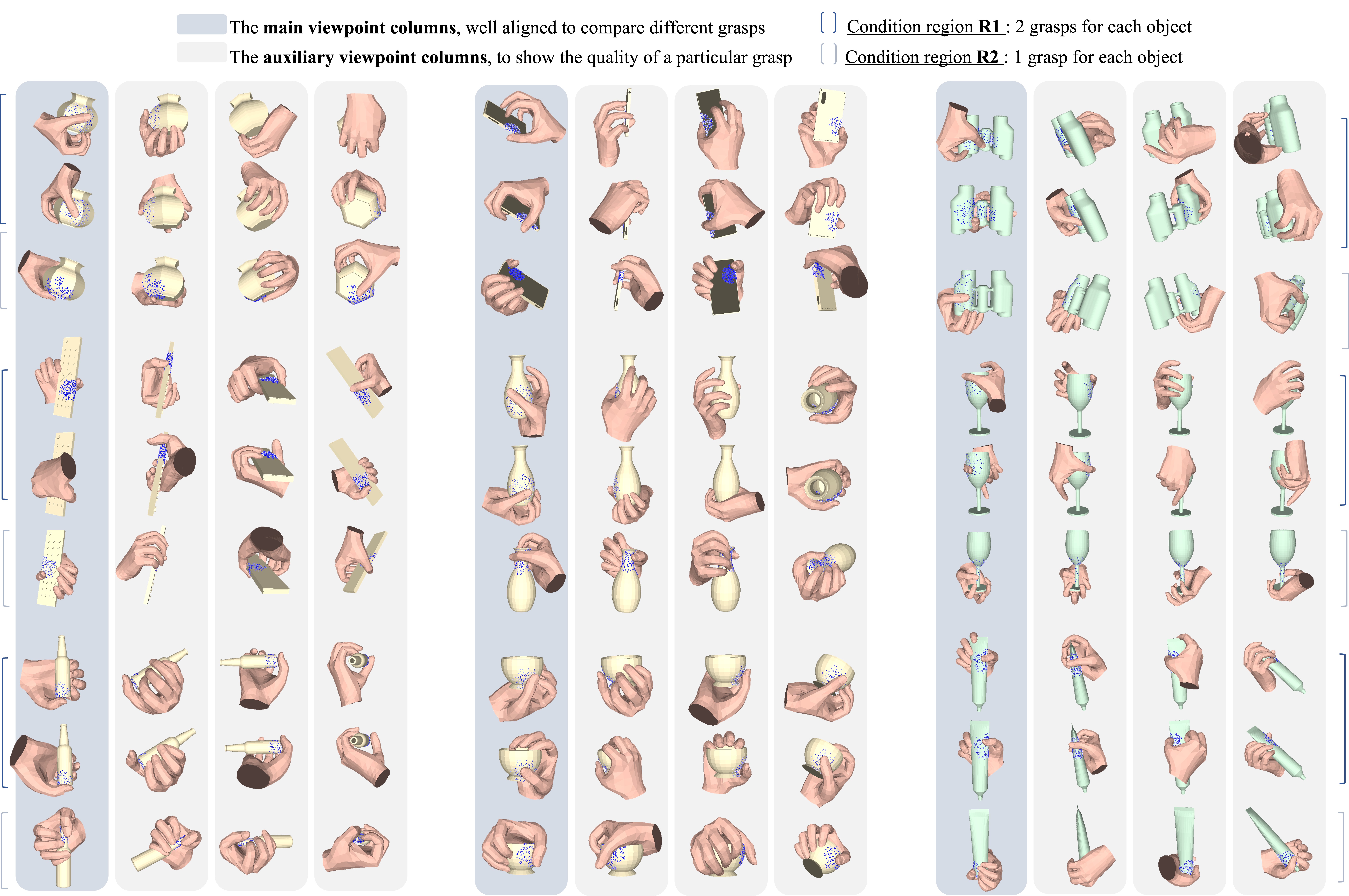}
\end{center}
   \caption{Generated hand grasps given different objects/condition regions from \textcolor{yellow}{ObMan\cite{hasson19_obman}(in domain)} and \textcolor{green}{GRAB\cite{GRAB}(out of domain)}. See supp. for more 3D demos.}
\label{fig:visual}
\end{figure*}

Figure \ref{fig:compare} shows the comparison of our method to GraspTTA\cite{graspTTA}, as it perform generally better than other methods compared. The visualization reasonably aligns with the quantitative result where \cite{graspTTA} presents low diversity and lack of controllability. Figure \ref{fig:visual} presents more visualization results of generated hand grasps given different objects, each with 2 different regions. Samples with green marked object presents the generalization results on out of domain objects from GrabNet testset. As visualized, the generated hand grasps are of great diversity, plausibility as well as accuracy of targeting the condition region with the thumb finger tip on both in and out of domain objects.

\subsubsection{User Study}
\begin{figure*}
\begin{center}
    \includegraphics[width=1.0\linewidth]{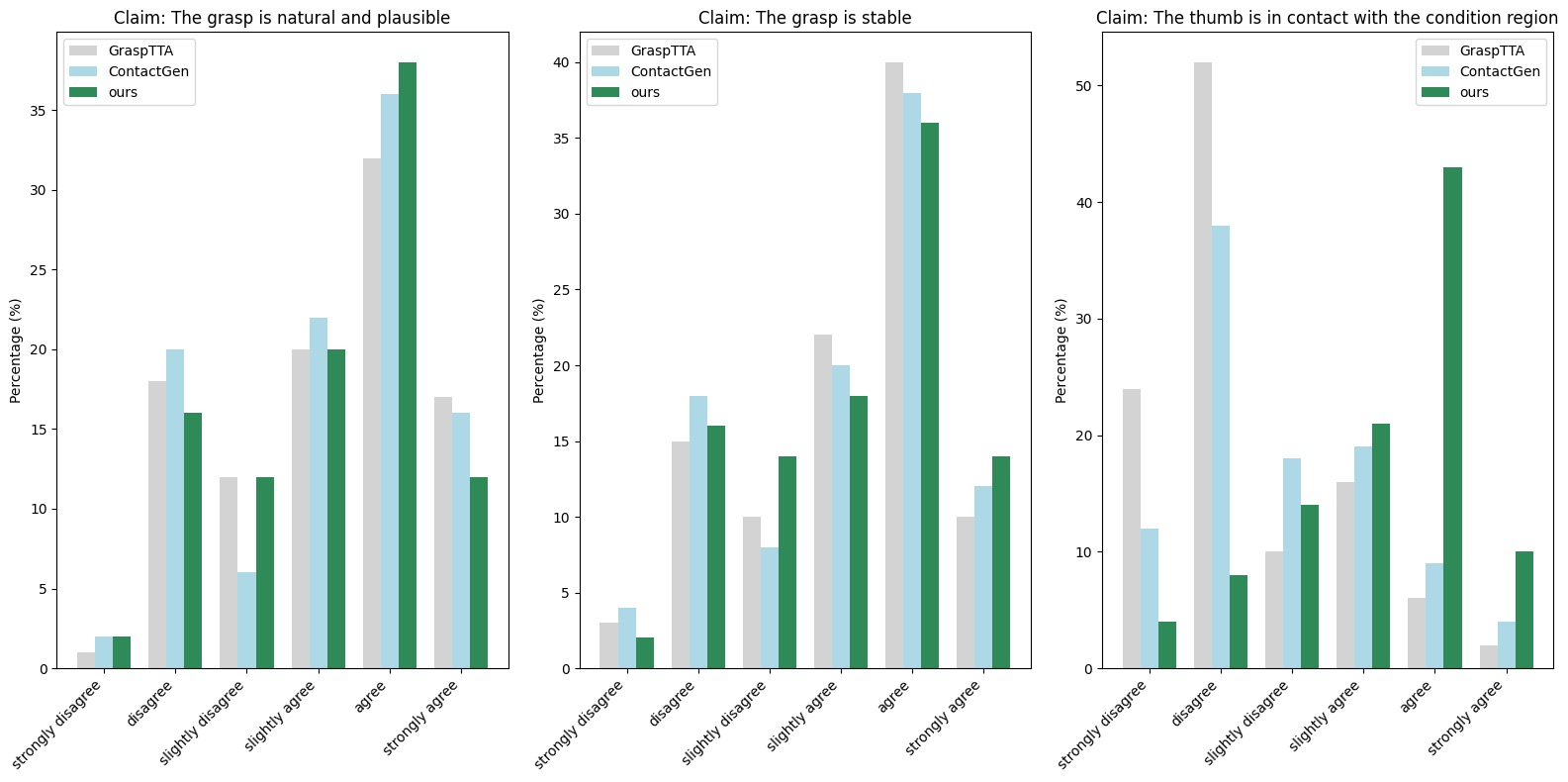}
\end{center}
   \caption{User study score distribution. The score distribution present the perceptual assessment of each method on the naturalness and plausibility, stability, and condition region controllability.}
\label{fig:user}
\end{figure*}

To assess the perceptual performances in our proposed RegionGrasp task, we conduct a user study among 20 participants comparing our method with GraspTTA\cite{graspTTA} and ContactGen\cite{liu2023contactgen}. We evaluate all the three methods on 10 objects in in-domain ObMan dataset and out-of-domain GRAB dataset. Each objects are provided with 5 randomly selected condition regions. The participants are asked to assess if the grasp looks 1) natural and plausible 2) stable 3) in contact with the condition region using the thumb, using 6 levels of scores from "strongly disagree" to "strongly agree". The results in Figure \ref{fig:user} display the score distribution of the three methods across all 6 assessment levels. As shown our method presents superior condition region controllability over the other two methods with comparable expressiveness in generating natural, plausible and stable hand grasps, which aligns with the evaluation results on quantitative metrics.


\begin{table}[tb]
  \caption{Ablation study on the effectiveness of the proposed ConditionNet, HOINet, and the pretrain strategy. 
  }
  \label{tab:ablation_model}
  \centering
  \begin{tabular}{@{}c c c c c c c @{}}
    \toprule
    { ConditionNet} & HOINet & Pretrain & CR Rate (\%)$\uparrow$ & CCA/IV$\uparrow$ & GDR$\downarrow$ & DivDist (mm)$\uparrow$ \\
    \midrule
      &  &  & 17.17 & 1.63 & 1.21 & 1.87 \\
    \checkmark &  &  & 75.37 & 0.88 & \textbf{1.10} & 2.21 \\
    \checkmark & \checkmark &  & 84.81 & 1.01 & 1.30 & 2.36 \\
    \checkmark & \checkmark & \checkmark & \textbf{87.88} & \textbf{1.73} & \textbf{1.10} & \textbf{2.48} \\
  \bottomrule
  \end{tabular}
\end{table}

\begin{table}[tb]
    \caption{Ablation study on the important parameters.}
    \label{tab:param}
    \centering
  \scalebox{1.0}{
  \begin{tabular}{@{}l c c c c@{}}
    \toprule
      & \textbf{CR Rate (\%)$\uparrow$} & \textbf{CCA/IV$\uparrow$} & \textbf{GDR$\downarrow$} & \textbf{DivDist (mm)$\uparrow$} \\
    \midrule
     $r_m = 20\%$ & 25.32 & 0.98 & 1.18 & 1.46 \\
     $r_m = 40\%$ & 71.59 & 1.32 & 1.20 & 2.21 \\
     $r_m = 80\%$ & 62.45 & 1.19 & 1.32 & \textbf{2.52} \\
     $r_m = 60\%$ & \textbf{87.88} & \textbf{1.63} & \textbf{1.10} & 2.48 \\
    \midrule
      $B_{hoi} = 1$ & 33.27 & 0.82 & 1.23 & 1.46 \\
      $B_{hoi} = 5$ & 79.25 & \textbf{1.63} & \textbf{1.07} & 0.76 \\
      $B_{hoi} = 3$ & \textbf{87.88} & \textbf{1.63} & 1.10 & \textbf{2.48} \\
    \bottomrule
  \end{tabular}
  }
\end{table}

\subsection{Ablation Study}

\subsubsection{Proposed Module Effectiveness}
Table \ref{tab:ablation_model} presents the results of the ablation study of each technical contribution. Row\#1 represents the baseline which adopts standard PointNet to encode hand and object separately, while row 5 represents our proposed model. We here mark \textbf{the best} for each metrics.As shown in Table \ref{tab:ablation_model}, great improvement in CR Rate can be observed by introducing ConditionNet. Similarly, combining HOINet with ConditionNet can also dramatically improve CR rate comparing. However, only introducing ConditionNet or with HOINet fails to result in satisfying hand grasp generation in general, as the overall grasp quality is inferior with either severe interpenetration problem or lack of grasp stability. By adopting pretraining strategy of randomly masking object point patches for O-Enc in ConditionNet, both the region controllability and overall grasp quality can be improved. The result demonstrated the synergistic effectiveness of our overall methodology.

\subsubsection{Model Parameter Setting}

We also present the ablation study on settings of important parameters: 1) the masking ratio $r_m$ of pretraining strategy 2)transformer blocks $B_{hoi}$ in the HOI encoder. As shown in Table \ref{tab:param}, masking ratio set either too high(80\%) or too low(20\%) limits the geometric-awareness of ConditionNet, resulting in much lower CR Rate and CCA/IV. Similarly, our model tends to generate limited hand grasp poses if designed with deep transformer layers for HOI encoder, as it overfits the hand-object interaction semantics during training.


\begin{figure*}
\begin{center}
    \includegraphics[width=0.5\linewidth]{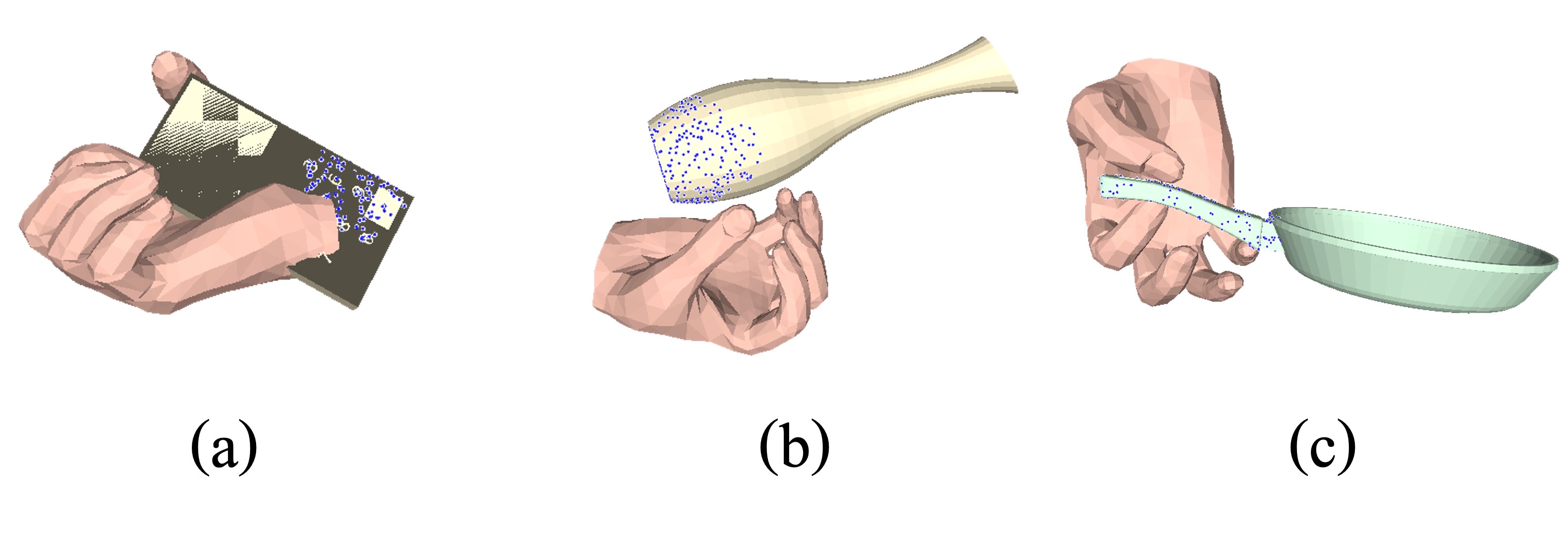}
\end{center}
   \caption{Failure cases. (a)Penetration in flat objects. (b)Lost of contact. (c)Lack of functional plausibility.}
\label{fig:failure}
\end{figure*}

\section{Limitations}

Despite the general satisfying results, we observed some failure cases. As shown in Figure \ref{fig:failure}(a), our model is less likely to succeed in flat objects. Sometimes generated hands like Figure \ref{fig:failure}(b) are targeting for condition region but not really in contact. Also in a few cases as Figure \ref{fig:failure}(c), hand grasps are not functionally plausible that lack contact clues. Prospectively, enhancing physical plausibility of grasps by integrating physic prior might a remedy. Meanwhile, incorporating Large Language Model(LLM) for annotating and analyzing the correlations between daily-base function of different objects and corresponding hand grasp poses is also a promising future path to take for a stronger hand-object prior. 
\section{Conclusion}

In this work, we introduce a novel task of region controllable hand grasp generation, facilitated by our proposed RegionGrasp-CVAE model. This model integrates ConditionNet for region-awareness and HOINet for interaction-awareness, effectively addressing the dual challenges of geometric and interaction complexities in hand grasp generation. Our empirical results validate the effectiveness of our approach, demonstrating significant advancements in generating diverse and plausible hand grasps. Given the limitation that failure cases occur in certain types of objects with lack of physical and functional plausibility, future works are motivated to focus on incorporating stronger prior based on physics simulation or foundation models. 
\bibliographystyle{splncs04}
\bibliography{main}

\begin{thebibliography}{10}
\providecommand{\url}[1]{\texttt{#1}}
\providecommand{\urlprefix}{URL }
\providecommand{\doi}[1]{https://doi.org/#1}

\bibitem{cha2024text2hoi}
Cha, J., Kim, J., Yoon, J.S., Baek, S.: Text2hoi: Text-guided 3d motion generation for hand-object interaction. In: Proceedings of the IEEE/CVF Conference on Computer Vision and Pattern Recognition (CVPR). pp. 1577--1585 (June 2024)

\bibitem{shapenet}
Chang, A.X., Funkhouser, T., Guibas, L., Hanrahan, P., Huang, Q., Li, Z., Savarese, S., Savva, M., Song, S., Su, H., et~al.: Shapenet: An information-rich 3d model repository. arXiv preprint arXiv:1512.03012  (2015)

\bibitem{chang2024text2grasp}
Chang, X., Sun, Y.: Text2grasp: Grasp synthesis by text prompts of object grasping parts. arXiv preprint arXiv:2404.15189  (2024)

\bibitem{christen2024diffh2o}
Christen, S., Hampali, S., Sener, F., Remelli, E., Hodan, T., Sauser, E., Ma, S., Tekin, B.: Diffh2o: Diffusion-based synthesis of hand-object interactions from textual descriptions. In: SIGGRAPH Asia 2024 (2024)

\bibitem{cvpr22dgrasp}
Christen, S., Kocabas, M., Aksan, E., Hwangbo, J., Song, J., Hilliges, O.: D-grasp: Physically plausible dynamic grasp synthesis for hand-object interactions. In: Proceedings of the IEEE/CVF Conference on Computer Vision and Pattern Recognition (CVPR) (2022)

\bibitem{contactopt}
Grady, P., Tang, C., Twigg, C.D., Vo, M., Brahmbhatt, S., Kemp, C.C.: {ContactOpt}: Optimizing contact to improve grasps. In: Conference on Computer Vision and Pattern Recognition (CVPR) (2021)

\bibitem{Guo_2022_CVPR}
Guo, C., Zou, S., Zuo, X., Wang, S., Ji, W., Li, X., Cheng, L.: Generating diverse and natural 3d human motions from text. In: Proceedings of the IEEE/CVF Conference on Computer Vision and Pattern Recognition (CVPR). pp. 5152--5161 (June 2022)

\bibitem{hasson19_obman}
Hasson, Y., Varol, G., Tzionas, D., Kalevatykh, I., Black, M.J., Laptev, I., Schmid, C.: Learning joint reconstruction of hands and manipulated objects. In: CVPR (2019)

\bibitem{graspTTA}
Jiang, H., Liu, S., Wang, J., Wang, X.: Hand-object contact consistency reasoning for human grasps generation. In: Proceedings of the International Conference on Computer Vision (2021)

\bibitem{graspfield}
Karunratanakul, K., Yang, J., Zhang, Y., Black, M.J., Muandet, K., Tang, S.: Grasping field: Learning implicit representations for human grasps. 2020 International Conference on 3D Vision (3DV) pp. 333--344 (2020)

\bibitem{contact2grasp}
Li, H., Lin, X., Zhou, Y., Li, X., Huo, Y., Chen, J., Ye, Q.: Contact2grasp: 3d grasp synthesis via hand-object contact constraint (2022). \doi{10.48550/ARXIV.2210.09245}, \url{https://arxiv.org/abs/2210.09245}

\bibitem{li2024graspmulti}
Li, Y., Liu, B., Geng, Y., Li, P., Yang, Y., Zhu, Y., Liu, T., Huang, S.: Grasp multiple objects with one hand. IEEE Robotics and Automation Letters  \textbf{9}(5),  1--10 (2024). \doi{10.1109/LRA.2024.1234567}

\bibitem{liu2023contactgen}
Liu, S., Zhou, Y., Yang, J., Gupta, S., Wang, S.: Contactgen: Generative contact modeling for grasp generation. In: Proceedings of the IEEE/CVF International Conference on Computer Vision (2023)

\bibitem{lundell2021multi}
Lundell, J., Corona, E., Le, T.N., Verdoja, F., Weinzaepfel, P., Rogez, G., Moreno-Noguer, F., Kyrki, V.: Multi-fingan: Generative coarse-to-fine sampling of multi-finger grasps. In: 2021 IEEE International Conference on Robotics and Automation (ICRA). pp. 4495--4501. IEEE (2021)

\bibitem{pointmae}
Pang, Y., Wang, W., Tay, F.E.H., Liu, W., Tian, Y., Yuan, L.: Masked autoencoders for point cloud self-supervised learning (2022)

\bibitem{pointnet}
Qi, C.R., Su, H., Mo, K., Guibas, L.J.: Pointnet: Deep learning on point sets for 3d classification and segmentation (2016), \url{http://arxiv.org/abs/1612.00593}, cite arxiv:1612.00593

\bibitem{pointnet++}
Qi, C.R., Yi, L., Su, H., Guibas, L.J.: Pointnet++: Deep hierarchical feature learning on point sets in a metric space (2017). \doi{10.48550/ARXIV.1706.02413}, \url{https://arxiv.org/abs/1706.02413}

\bibitem{CVAE}
Qi, C.R., Yi, L., Su, H., Guibas, L.J.: Pointnet++: Deep hierarchical feature learning on point sets in a metric space. In: Guyon, I., Luxburg, U.V., Bengio, S., Wallach, H., Fergus, R., Vishwanathan, S., Garnett, R. (eds.) Advances in Neural Information Processing Systems. vol.~30. Curran Associates, Inc. (2017), \url{https://proceedings.neurips.cc/paper/2017/file/d8bf84be3800d12f74d8b05e9b89836f-Paper.pdf}

\bibitem{MANO:SIGGRAPHASIA:2017}
Romero, J., Tzionas, D., Black, M.J.: Embodied hands: Modeling and capturing hands and bodies together. ACM Transactions on Graphics, (Proc. SIGGRAPH Asia)  \textbf{36}(6) (Nov 2017)

\bibitem{GRAB}
Taheri, O., Ghorbani, N., Black, M.J., Tzionas, D.: {GRAB}: A dataset of whole-body human grasping of objects. In: ECCV (2020), \url{https://grab.is.tue.mpg.de}

\bibitem{eccv22graspD}
Turpin, D., Wang, L., Heiden, E., Chen, Y.C., Macklin, M., Tsogkas, S., Dickinson, S., Garg, A.: Grasp’D: Differentiable Contact-Rich Grasp Synthesis for Multi-Fingered Hands, pp. 201--221 (11 2022). \doi{10.1007/978-3-031-20068-7_12}

\bibitem{turpin2023fastgraspd}
Turpin, D., Zhong, T., Zhang, S., Zhu, G., Heiden, E., Macklin, M., Tsogkas, S., Dickinson, S., Garg, A.: Fast-grasp'd: Dexterous multi-finger grasp generation through differentiable simulation. In: ICRA (2023)

\bibitem{xu2024dgtr}
Xu, G.H., Wei, Y.L., Zheng, D., Wu, X.M., Zheng, W.S.: Dexterous grasp transformer. In: Proceedings of the IEEE/CVF Conference on Computer Vision and Pattern Recognition (CVPR). pp. 17933--17942 (June 2024)

\bibitem{xu2023unidexgrasp}
Xu, Y., Wan, W., Zhang, J., Liu, H., Shan, Z., Shen, H., Wang, R., Geng, H., Weng, Y., Chen, J., et~al.: Unidexgrasp: Universal robotic dexterous grasping via learning diverse proposal generation and goal-conditioned policy. arXiv preprint arXiv:2303.00938  (2023)

\bibitem{ye2023ghop}
Ye, Y., Gupta, A., Kitani, K., Tulsiani, S.: G-hop: Generative hand-object prior for interaction reconstruction and grasp synthesis. In: CVPR (2024)

\bibitem{yu2021pointr}
Yu, X., Rao, Y., Wang, Z., Liu, Z., Lu, J., Zhou, J.: Pointr: Diverse point cloud completion with geometry-aware transformers. In: ICCV (2021)

\bibitem{pointbert}
Yu, X., Tang, L., Rao, Y., Huang, T., Zhou, J., Lu, J.: Point-bert: Pre-training 3d point cloud transformers with masked point modeling. In: Proceedings of the IEEE Conference on Computer Vision and Pattern Recognition (CVPR) (2022)

\bibitem{yuan2018pcn}
Yuan, W., Khot, T., Held, D., Mertz, C., Hebert, M.: Pcn: Point completion network. In: 3D Vision (3DV), 2018 International Conference on (2018)

\bibitem{pointtrans}
Zhao, H., Jiang, L., Jia, J., Torr, P.H., Koltun, V.: Point transformer. In: Proceedings of the IEEE/CVF International Conference on Computer Vision. pp. 16259--16268 (2021)

\bibitem{zhou2022toch}
Zhou, K., Bhatnagar, B.L., Lenssen, J.E., Pons-Moll, G.: Toch: Spatio-temporal object correspondence to hand for motion refinement. In: European Conference on Computer Vision ({ECCV}). {Springer} (October 2022)

\end{thebibliography}
\end{document}